\documentclass{article}
\usepackage{amsmath,epsfig}
\usepackage[preprint]{spconfa4}
\usepackage{multirow, subfigure, amssymb}

\copyrightnotice{\textbf{978-1-6654-3864-3/21/\$31.00~\copyright 2021 IEEE}}

\let\OLDthebibliography\thebibliography
\renewcommand\thebibliography[1]{
  \OLDthebibliography{#1}
  \setlength{\parskip}{0pt}
  \setlength{\itemsep}{0pt plus 0.3ex}
}

\begin{document}\sloppy

\def\x{{\mathbf x}}
\def\L{{\cal L}}

\title{Deep Feature Selection-and-Fusion for RGB-D Semantic Segmentation}
%
\name{Yuejiao Su, Yuan Yuan, Zhiyu Jiang$^{\ast}$
\thanks{2021 IEEE. Personal use of this material is permitted. Permission from IEEE must be obtained for all other uses, in any current or future media, including reprinting/republishing this material for advertising or promotional purposes, creating new collective works, for resale or redistribution to servers or lists, or reuse of any copyrighted component of this work in other works.. *Corresponding author: Zhiyu Jiang (jiangzhiyu@nwpu.edu.cn)}}

\address{School of Computer Science and School of Artificial Intelligence, Optics and Electronics (iOPEN),\\
Northwestern Polytechnical University, Xi'an 710072, P.R. China\\
yuejiao@mail.nwpu.edu.cn; y.yuan1.ieee@gmail.com; jiangzhiyu@nwpu.edu.cn}


\maketitle

\begin{abstract}
Scene depth information can help visual information for more accurate semantic segmentation. However, how to effectively integrate multi-modality information into representative features is still an open problem. Most of the existing work uses DCNNs to implicitly fuse multi-modality information. But as the network deepens, some critical distinguishing features may be lost, which reduces the segmentation performance. This work proposes a unified and efficient feature selection-and-fusion network (FSFNet), which contains a symmetric cross-modality residual fusion module used for explicit fusion of multi-modality information. Besides, the network includes a detailed feature propagation module, which is used to maintain low-level detailed information during the forward process of the network. Compared with the state-of-the-art methods, experimental evaluations demonstrate that the proposed model achieves competitive performance on two public datasets.
\end{abstract}
\begin{keywords}
RGB-D Semantic Segmentation, Multi-modality, Skip-connection, Attention Mechanism
\end{keywords}
\section{Introduction}
\label{sec:intro}

Semantic segmentation refers to the pixel-wise classification of images according to semantic information. Besides widely-used visual information, the depth information is regarded as another supplementary information to improve the scenario understanding performance due to the development of depth sensors. Depth modality contains 3D geometric information, which is insensitive to illumination changes and can distinguish objects with similar appearance. Therefore, depth cues can make up for some of the defects of semantic segmentation using only visual cues. RGB-D semantic segmentation is very important for many applications such as autonomous driving~\cite{DBLP:conf/icmcs/YangBD20}, robot vision and understanding~\cite{DBLP:conf/iros/MaSKC17}, and land cover classification~\cite{DBLP:conf/icmcs/LiuHL20}, \emph{etc}.

With the development of deep learning, two-stream networks have achieved remarkable performance in RGB-D semantic segmentation~\cite{DBLP:conf/cvpr/ChengCLZH17, DBLP:conf/eccv/GuptaGAM14, DBLP:conf/cvpr/Xiong0G020}. As we all know, the information of RGB and depth modalities are complementary. However, how to effectively fuse RGB and depth information into a unified and distinguishing representation is still a basic yet difficult issue in RGB-D semantic segmentation. There are many methods to try to solve this problem. FuseNet~\cite{DBLP:conf/accv/HazirbasMDC16} and RedNet~\cite{DBLP:journals/corr/abs-1806-01054} integrated different modalities by directly adding the feature maps of depth to RGB. RFBNet~\cite{DBLP:journals/corr/abs-1907-00135} proposed a residual fusion block to achieve bottom-up interaction and fusion between two modalities. ACNet~\cite{DBLP:conf/icip/HuYFW19} proposed an attention complementary module to assign different modality-weights for better integration. RDFNet~\cite{DBLP:conf/iccv/LeePH17} used single-modality residual learning to learn residual RGB and depth features and their combinations to exploit the complementary characteristics. Although these methods provide structured models to integrate the two kinds of information, it is still an unresolved problem to ensure that the network makes full use of the information from both modalities for fine semantic segmentation.

Moreover, the loss of detailed information during down-sampling is an inherent property of convolution operation. For the encoding part of the segmentation network, reducing the resolution of the feature map to a very small level through various pooling layers is not conducive to accurate mask generation, which can lead to inaccurate segmentation results. To further make up for the information lost in the encoding stage, U-Net~\cite{DBLP:conf/miccai/RonnebergerFB15} proposed skip-connection to reuse the feature to assist up-sampling learning and recover the fine segmentation results. Although it can realize the reuse of some lost features, it lacks pertinence and does not explicitly model the recovery of detailed information.

To solve the above problems, this work proposes a novel feature selection-and-fusion network to explicitly strengthen features in RGB-D semantic segmentation model from two aspects: multi-modality representations and decoder features. The key idea of our proposed network is to select discriminative information from one modality to supplement the other modality to obtain well-informed representation. In addition, this work focuses on the lost information in the encoder and finds ways to make it helpful in predicting the final result. These two aspects correspond to two modules respectively. For the former, the Symmetric Cross-modality Residual Fusion module (SCRF) is designed to effectively fuse the complementary information of two modalities, while maintaining the specificity of the specific modality during the information interaction process at the encoder stage. For the latter, a Detailed Feature Propagation module (DFP) is designed to encourage the network to spotlight the missing vital details in the encoder and reuse them in the decoder to improve the segmentation performance. Both modules are designed as two steps: feature selection and feature fusion.

The main contributions of this work are described as follows:
\begin{itemize}
  \item In order to solve the multi-modality information fusion in RGB-D semantic segmentation, this work designs the SCRF module in FSFNet. The core of the module is the cross-modality residual connection, which can retain the advantages of the residual connection and can explicitly select and fuse complementary information into distinguishing and effective representations.
  \item In response to the loss of some important information during the down-sampling process, this work designs the DFP module in the network. The DFP module firstly selects vital information that may be lost in the encoder stage by attention mechanism. And then the module propagates and fuse the selected features with decoder features for further segmentation.
  \item With proposed modules, our FSFNet uses a relatively simple architecture to achieve excellent performance. We verify the effectiveness of FSFNet and its modules through a series of experiments and achieve competitive or superior performance on NYUDv2 and SUN RGB-D datasets.
\end{itemize}

\section{Related Work}
The main difference between RGB-D semantic segmentation and RGB semantic segmentation is that the former not only use visual information but also leverages scene depth information to achieve better accuracy~\cite{DBLP:journals/ijon/JiangYW18}. According to whether the deep learning methods are used or not, RGB-D semantic segmentation methods are roughly divided into traditional and deep learning-based methods.

In the early stage, researchers preferred to use depth information directly. Silberman \emph{et al}.~\cite{DBLP:conf/eccv/SilbermanHKF12} recovered support relationships by parsing indoor scenes into floor, walls, supporting surfaces, and object regions using depth information. Ren \emph{et al}.~\cite{DBLP:conf/cvpr/RenBF12} achieved high labeling accuracy by a combination of color and depth features using kernel descriptors, and by combining MRF with segmentation tree. Gupta \emph{et al}.~\cite{DBLP:conf/cvpr/GuptaAM13} proposed algorithms for object boundary detection and hierarchical segmentation that generalize the gPb-ucm~\cite{DBLP:journals/pami/ArbelaezMFM11} approach by making effective use of depth information.

Although the traditional methods can mine the internal relationship between RGB and depth information by explicitly utilizing depth information, they need a lot of prior knowledge and specific descriptors. Compared with traditional methods, deep learning-based approaches implicitly utilizes deep information to assist RGB semantic segmentation through various networks. Couprie \emph{et al}.~\cite{DBLP:journals/corr/abs-1301-3572,DBLP:conf/eccv/WangWTSW16} regarded the depth information as another channel to concatenate with RGB image and then extract features with RGB semantic segmentation network. However, RGB channel and depth channel contain inconsistent features and cannot be processed by shared network feature extractors. Besides, Gupta \emph{et al}.~\cite{DBLP:conf/eccv/GuptaGAM14} encoded depth into HHA (Horizontal disparity, Height above ground, and Angle with gravity). Some studies~\cite{DBLP:journals/corr/abs-1806-01054,DBLP:journals/corr/abs-1907-00135,DBLP:conf/icip/HuYFW19,DBLP:journals/corr/abs-2007-09183} used two-stream networks to process the RGB images and HHA information. These methods prove that depth data can improve the performance of semantic segmentation. Many of the later researchers focus on the effective fusion of multi-modality data. Proposed by Hazirbas \emph{et al}.~\cite{DBLP:conf/accv/HazirbasMDC16}, FuseNet integrated depth characteristics into RGB feature mapping by dense fusion or sparse fusion strategies as the network deepens. Lin \emph{et al}.~\cite{DBLP:conf/iccv/LinCCHH17} divided the image into several branches with different scene resolutions based on depth information aiming at the multi-scale problem. Compared with single-scale networks, multi-scale networks~\cite{DBLP:conf/icmcs/YuanJW17} have better segmentation performance, but they also require more computation. Different from the previous work, this work extends the idea of residual connection~\cite{DBLP:conf/cvpr/HeZRS16} to multi-modality and designs a SCRF module based on multi-modal residual connection to clearly promote multi-modality feature fusion.
\begin{figure*}[ht]
 \centering
 \includegraphics[scale=0.6]{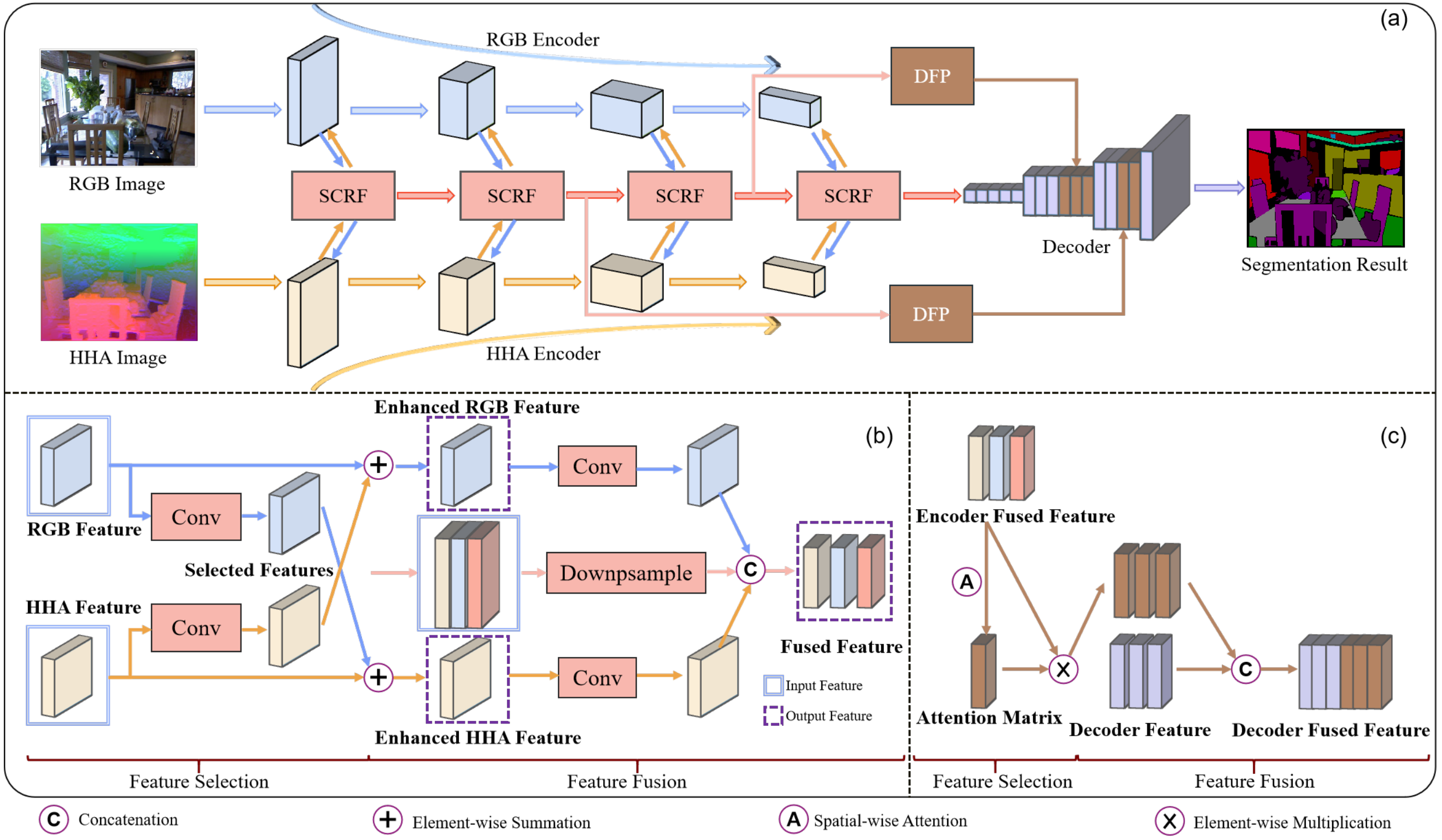}
 \caption{(a) Overview of the proposed framework. Given RGB and HHA images as input, the upper and lower encoder branches extract the characteristics of specific modalities from input respectively. The middle fusion branch uses cascaded SCRF modules to fuse the two modal features. The fused features of the middle two layers of the fusion branch are selected by the DFP module and propagated to the corresponding decoder layer for joint prediction. (b) Details of SCRF module. It is based on the cross-modality residual connection. The SCRF firstly select features that complement another modality from one modality, and then perform feature fusion between modalities and levels. (c) Details of DFP module. The DFP firstly uses spatial-wise attention to select important but possibly lost detailed information from the fusion features in the middle two layers of the encoder stage, and then merge them with the corresponding features of the decoder stage for the final joint segmentation.}
 \label{framework}

\end{figure*}

\section{PROPOSED METHOD}

\subsection{Overview}

The RGB-D semantic segmentation model is usually based on the encoder-decoder architecture. However, a superior encoder can unify both the complementary characteristics of two modalities and their specific characteristics into effective representation. To this end, inspired by the residual connection~\cite{DBLP:conf/cvpr/HeZRS16}, we put forward a unified encoder framework that includes the symmetric cross-modality residual fusion module as described in Section~\ref{scrf}, which aims to encourage explicit fusion of cross-modality information and preserve single-modality specific characteristics as completely as possible. In addition, in the encoder phase, a lot of detailed information is lost due to cascading downsampling, which is fatal to semantic segmentation. Our goal is to automatically reuse the essential details lost in the encoder stage and make them helpful to form the final segmentation mask. For this purpose, the detailed feature propagation module is proposed between encoder and decoder, as described in Section~\ref{dfp}.

The overall frame diagram of the proposed method is shown in Fig.~\ref{framework}(a). Take the three-channel RGB image $\textbf{I}_{\text{R}} \in \mathbb{R}^{H \times W \times 3}$ and three-channel HHA image $\textbf{I}_{\text{H}} \in \mathbb{R}^{\emph{H} \times \emph{W} \times 3}$ as input, our network selects and enhances the information representation capability of two modalities through cascaded SCRF modules, and at the same time encourages to save the specific features of the specific modality as much as possible. In addition, partial details of the encoder are selected and transmitted by the DFP module to the corresponding stage of the decoder to make full use of the lost important details. Each component will be described in more details in the remaining parts of this section.

\subsection{Symmetric Cross-modality Residual Fusion}

\label{scrf}
To encourage the complementary fusion of information between two modalities, we consider this issue from two aspects. Firstly, we argue that if the complementary part can be modeled more explicitly, cross-modality information fusion can be accomplished better. For this motivation, the SCRF module in every layer is designed into two steps, feature selection and feature fusion, as shown in Fig.~\ref{framework}(b). Unlike other work using the simple single-modality residual connection~\cite{DBLP:conf/iccv/LeePH17}, we propose a cross-modality residual connection to select and fuse complementary features from another modality to make both steps effectively. Secondly, as we know, shallow features can identify edge information, while deep features can study the global context to locate salient objects~\cite{DBLP:conf/cvpr/ZhaoW19}. Therefore, the deep features are very irregular, while the shallow features are very noisy and chaotic, and a lot of previous work lacked cross-level interaction in the fusion branch. So in this paper, each SCRF module of the fusion branch is cascaded from shallow to deep, so that the network can gradually select complementary features across levels for joint decision-making. Assume that the RGB feature map of $\emph{j}$-th layer is $\textbf{F}_{\text{rgb}}^{j} \in \mathbb{R}^{\emph{H} \times \emph{W} \times \emph{C}}$, and the HHA feature map of $\emph{j}$-th layer is $\textbf{F}_{\text{hha}}^{j} \in \mathbb{R}^{\emph{H} \times \emph{W} \times \emph{C}}$. For simplicity, $\emph{H}$, $\emph{W}$, and $\emph{C}$ are the height, width, and channel of the feature map respectively. So the selection process can be denoted as:

\begin{equation}
\textbf{S}_{\text{hha}}^\emph{j} = \emph{f}_{{\text{s}_1}}^{}(\textbf{F}_{\text{hha}}^\emph{j}),\textbf{S}_{\text{rgb}}^\emph{j} = \emph{f}_{{\text{s}_2}}^{}(\textbf{F}_{\text{rgb}}^\emph{j}),
\end{equation}
where $\textbf{S}_{\text{rgb}}^j$ and $\textbf{S}_{\text{hha}}^j$ are selected RGB and HHA features of $\emph{j}$-th layer. They are automatically selected and they can help improve the distinguishing ability of another modality. Note that $\emph{f}_{{\text{s}_1}}^{}( \cdot )$ and $\emph{f}_{{\text{s}_2}}^{}( \cdot )$ are feature selection operations. In practice, we use $1 \times 1$ convolution with stride 1.

And then the selected features are sent to the feature fusion step. Suppose that the fusion feature of the previous layer of $\emph{j}$-th layer is $\textbf{F}_{\text{fuse}}^{j-1} \in \mathbb{R}^{\emph{H} \times \emph{W} \times \emph{C}}$. Then the feature after fusion is:
\begin{small}
\begin{equation}
\begin{split}
\label{qu2}
\textbf{F}_{\text{fuse}}^\emph{j} = [{\emph{f}_{\text{down}}}(\textbf{F}_{\text{fuse}}^{\emph{j} - 1}),{\emph{f}_{\text{conv}}}(\textbf{S}_{\text{hha}}^\emph{j} +\\
 \textbf{F}_{\text{rgb}}^\emph{j}),
{\emph{f}_{\text{conv}}}(\textbf{S}_{\text{rgb}}^\emph{j} + \textbf{F}_{\text{hha}}^\emph{j})].
\end{split}
\end{equation}
\end{small}

Here $\textbf{F}_{\text{fuse}}^\emph{j}$ is the concatenated fused feature. ${\emph{f}_{\text{down}}}( \cdot )$ and ${\emph{f}_{\text{conv}}}( \cdot )$ mean the down-sample and convolution operation respectively. Note that when $\emph{j}=1$, only the last two items in equation (\ref{qu2}) are concatenated.

Therefore cross-modality residual function we defined includes feature selection, feature fusion, and the final convolution part, that is:
\begin{equation}
\begin{array}{l}
\emph{f}_{{r_1}}^{}:\textbf{F}_{\text{rgb}}^\emph{j} \to {\emph{f}_{\text{conv}}}(\textbf{S}_{\text{hha}}^\emph{j} + \textbf{F}_{\text{rgb}}^\emph{j}),\\
\emph{f}_{r_2}^{}:\textbf{F}_{\text{hha}}^\emph{j} \to {\emph{f}_{\text{conv}}}(\textbf{S}_{\text{rgb}}^\emph{j} + \textbf{F}_{\text{hha}}^\emph{j}),
\end{array}
\end{equation}
where $\emph{f}_{{r_1}}^{}( \cdot )$ and $\emph{f}_{r_2}^{}( \cdot )$ are the cross-modality residual functions we defined.

The simple but effective SCRF module not only continues the advantages of residual connection, that is, it can accelerate the convergence of the network, but also it uses adaptive weight learning to intelligently and explicitly select and fuse supplementary features from another modality. And our work enables the network to automatically select deep or shallow features by cascading the module. These all allow us to effectively aggregate multi-modality information.

\subsection{Detailed Feature Propagation}

\label{dfp}
As mentioned above, the semantic information of some small-scale objects and outlines will be lost in the down-sampling stage, which is unfavorable for semantic segmentation. Therefore, inspired by the idea of skip-connection~\cite{DBLP:conf/miccai/RonnebergerFB15}, we design the detailed feature propagation module to reuse the lost but essential information, as shown in Fig.~\ref{framework}(c).

The DFP module is still designed with the same two steps as the SCRF module, \emph{i.e.} feature selection and feature fusion. Different from SCRF, DFP automatically selects some detailed features that may be lost in the encoder stage by spatial-wise attention, and then propagates the selected features to the corresponding decoder part for fusion with the features being up-sampled. In this way, the features of some small scale objects and outlines can be well used.

For feature selection, we make the model select detailed features only from the second and third layers of the encoder. This is because the first layer contains a lot of noise and is very complex, and the features of the fourth layer no longer contain enough detailed features after the cascading of the down-sampling operations.

Accordingly, our feature fusion operation is carried out in the corresponding decoder layer, to achieve the enhancement of features in the decoder, which can improve the segmentation ability of some small objects and outlines. Suppose fused feature maps in $\emph{i}$-th layer of encoder denoted as $\textbf{F}_{\text{encoder}}^\emph{i} \in \mathbb{R}^{\emph{H} \times \emph{W} \times \emph{C}}$, the feature maps of the corresponding decoder are $\textbf{F}_{\text{decoder}}^\emph{m-i} \in \mathbb{R}^{\emph{H} \times \emph{W} \times \emph{C}}$, then the enhanced decoder features after feature selection-and-fusion can be calculated by:
\begin{equation}
\small
\tilde{\textbf{F}}_{\text{decoder}}^\emph{m-i} = {\emph{f}_{\text{fuse}}}({\emph{f}_{\text{select}}}(\textbf{F}_{\text{encoder}}^\emph{i}),\textbf{F}_{\text{decoder}}^\emph{m-i}), \emph{i} \in\{2,3\}, \emph{m}=4.
\end{equation}

Here, $\tilde{\textbf{F}}_{\text{decoder}}^\emph{m-i}$ represents the feature maps after fusion, $\emph{i} \in\{2,3\}$ means that the DFP module only works in the second and third layers, while ${\emph{f}_{\text{select}}}( \cdot )$ and ${\emph{f}_{\text{fuse}}}( \cdot )$ represents feature selection operation and feature fusion operation, respectively.
In practice, we use the spatial-wise attention as the feature selection block to select some useful but may be lost features, and we transfer the processed features to the corresponding decoder layer like skip-connection. Simple concatenation is used as the feature fusion operation.

Through explicitly selecting and fusing detailed features, the impact of the loss of detailed features in down-sampling on the final segmentation can be reduced. And the semantic information of some small-scale objects and outlines can be preserved to realize the reuse of features, which can play their role in the prediction of the final segmentation mask.

In addition, we use pyramid supervision~\cite{DBLP:conf/icip/HuYFW19} in our work. In other words, the output of the last three layers of the decoder is also used as supervision information to ensure rapid network convergence. The only difference between the supervision information of the intermediate output and the ground truth is the resolution. We obtain the supervision information of the intermediate output through the nearest neighbor interpolation down-sampling. The final loss function is as follows:
\begin{equation}
{L_{{\rm{final}}}}{\rm{ = }}\sum\limits_{\emph{i} = 1}^3 {{\lambda _\emph{i}}} l_\emph{i}^{},
\end{equation}
where $\emph{l}_{\emph{i}}$ and $\lambda_{\emph{i}}$ represents the loss function and its weight of the $\emph{i}$-th layer, respectively. The weighted cross-entropy loss function is the loss function of each layer. Our hyper-parameter settings refer to ACNet~\cite{DBLP:conf/icip/HuYFW19}.
\begin{figure*}[ht]
 \centering
 \includegraphics[scale=0.35]{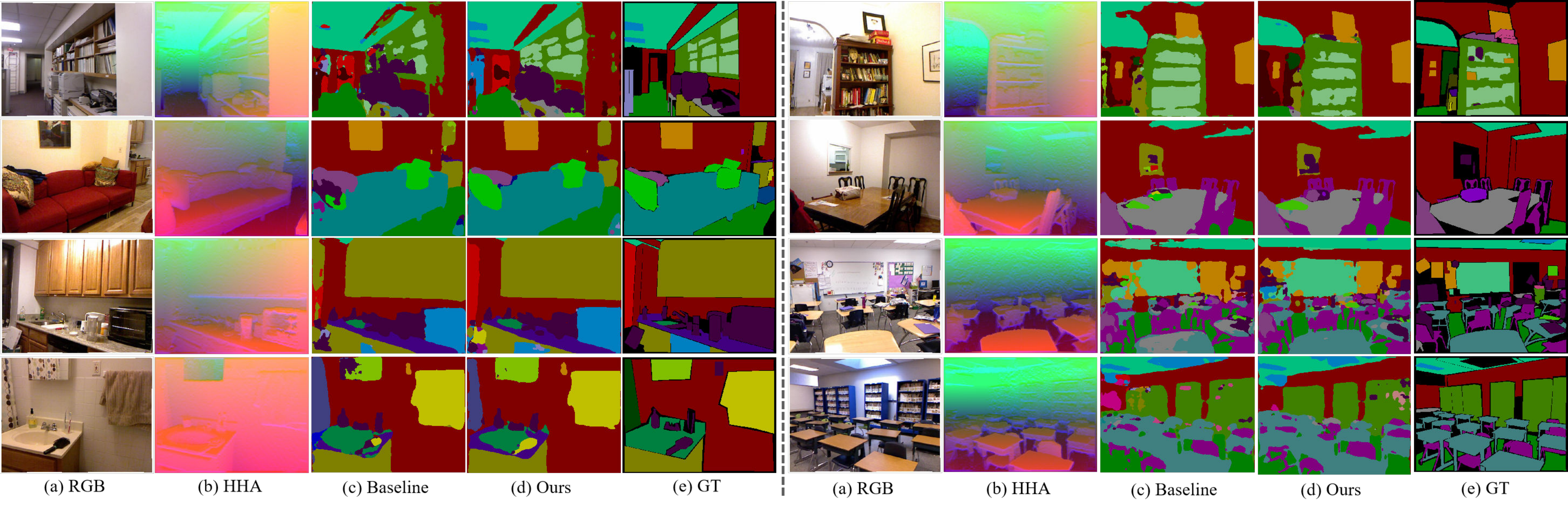}
 \caption{Visualization results of this work in NYUDv2 (left) and SUN RGB-D (right) test sets. For each dataset, we show (a) RGB image, (b) HHA image, (c) result of baseline, (d) result of ours, and (e) ground truth. The baseline directly adds feature maps in HHA branch into RGB branch.}
 \label{result}
 \vspace{-0.3cm}
\end{figure*}

\section{experiments}

\subsection{Datasets and Evaluation Metrics}

NYUDv2~\cite{DBLP:conf/eccv/SilbermanHKF12} and SUN RGB-D~\cite{DBLP:conf/cvpr/SongLX15} datasets are used to evaluate the proposed network. NYUDv2 dataset contains 1449 densely labeled pairs of RGB-D images captured by Microsoft Kinect. There are 795 pairs of images for training and 654 pairs for testing. The SUN RGB-D dataset is the largest RGB-D semantic segmentation dataset currently, with 10,335 densely annotated RGB-D images taken from 20 different scenes. It is captured by four different sensors (Kinect V1, Kinect V2, Xtion, and RealSense). The officially divided training set consists of 5285 pairs of RGB-D images and labels, and the remaining 5050 pairs are used for testing. The number of classes in both datasets is 40.

As with most related work, mean Intersection-over-Union (mIoU) and Pixel Accuracy (Pixel Acc.) are used as our performance metrics.

\subsection{Implementation Details}

PyTorch framework is used in our work and we use Stochastic Gradient Descent (SGD) optimizer to train our model with a momentum of 0.9 and a weight decay of 0.0005. Initial learning rate in our work is set to 0.02 and decreased at a rate of 0.9. The input RGB-D images are cropped to ${\rm{480}} \times {\rm{480}}$, and the batch size is set to 10.

\subsection{Experimental Results}

To prove the effectiveness of the proposed model, we compare it with several state-of-the-art methods, as shown in Table~\ref{overall}. The result shows that on NYUDv2 dataset, our FSFNet performs equivalently to the current best model using more simple architecture. And on SUN RGB-D dataset, our FSFNet outperforms other models in mIoU. That's because our model uses cascaded SCRF modules to implicitly integrate multi-modality features and uses the DFP module to reuse the selected detailed information from the encoder, which improves the ability to segment small-scale objects and contours.
\begin{table}[t]
\small
\begin{center}
\caption{Comparing with state-of-the-art methods on NYUDv2 and SUN RGB-D test sets.}
\label{overall}
\begin{tabular}{lcccc}
\hline
\multirow{2}{*}{Method} & \multicolumn{2}{c}{NYUDv2}        & \multicolumn{2}{c}{SUN RGB-D}     \\  \cline{2-5}
                         & mIoU            & Pixel Acc.      & mIoU            & Pixel Acc.      \\ \hline
3DGNN~\cite{DBLP:conf/iccv/QiLJFU17}                    & 43.1\%          & -               & 45.9\%          & -               \\
Kong \emph{et al}.~\cite{DBLP:conf/cvpr/KongF18}              & 44.5\%          & 72.1\%          & 45.1\%          & 80.3\%          \\
RedNet~\cite{DBLP:journals/corr/abs-1806-01054}                   & -               & -               & 47.8\%          & 81.3\%          \\
CFN~\cite{DBLP:conf/iccv/LinCCHH17}                      & 47.7\%          & -               & 48.1\%          & -               \\
ACNet~\cite{DBLP:conf/icip/HuYFW19}                    & 48.3\%          & -               & 48.1\%          & -               \\
PAP~\cite{DBLP:conf/cvpr/0005CXYS019}                      & 50.4\%          & 76.2\%          & 50.5\%          & \textbf{83.8\%} \\
SA-Gate~\cite{DBLP:journals/corr/abs-2007-09183}         & \textbf{52.4\%} & \textbf{77.9\%}          & 49.4\%          & 82.5\%          \\ \hline
\textbf{Ours}            & 52.0\%          & \textbf{77.9\%} & \textbf{50.6\%} & 81.8\%          \\ \hline
\end{tabular}
\end{center}
\vspace{-0.4cm}
\end{table}

In order to verify the effectiveness of the SCRF and DFP modules, we perform ablation studies on the NYUDv2 dataset under the same parameter settings as shown in Table~\ref{overall}. We use ResNet-101 as our backbone and the first row in Table~\ref{ablation} is the baseline that fuses RGB and depth feature maps by element-wise summation in each encoder layer. We can observe that SCRF and DFP can improve performance by $2.9\%$ and $1.7\%$ respectively. And when two modules exist at the same time, the performance improvement is more obvious than the baseline. This experiment proves the effectiveness and importance of the two proposed modules.

\begin{table}[t]
\begin{center}
\caption{Ablation study for SCRF and DFP modules on NYUDv2 dataset.}
\label{ablation}
\begin{tabular}{lcl}
\cline{1-2}
Method              & mIoU(\%)             &  \\ \cline{1-2}
Res-101 + SUM        & 47.9                 &  \\
Res-101 + DFP        & 49.6(1.7\%$\uparrow$)          &  \\
Res-101 + SCRF       & 50.8(2.9\%$\uparrow$)          &  \\
Res-101 + SCRF + DFP & \textbf{52.0(4.1\%$\uparrow$)} &  \\ \cline{1-2}
\end{tabular}
\end{center}
\vspace{-0.4cm}
\end{table}

In order to display the results of the model more intuitively, we show a part of the visualization results in Fig.~\ref{result}. It can be observed that compared to the baseline, our model has a better segmentation performance on different classes of objects, such as ceiling, table, \emph{etc.}, which can prove that our model combines the characteristics of RGB and depth information well. In addition, our model can distinguish small-scale objects and can perform more accurate contour segmentation, which can prove the effectiveness of DFP.

\section{conclusions}

In this work, we propose a feature selection-and-fusion network to address two main challenges in RGB-D semantic segmentation. Firstly, for the effective fusion of multi-modality information, we put forward the cascaded SCRF module to obtain unified multi-modality representations. Secondly, aimed at the loss of detailed information in the down-sampling stage, we design the DFP module to make the important details helpful in predicting the results. Experimental results demonstrate our model achieves competitive performance on NYUDv2 and SUN RGB-D datasets.

\footnotesize
\bibliographystyle{IEEEbib}
\bibliography{icme2021template}

\end{document}